%% file: main.tex
\documentclass[10pt]{article}

\input{hicss-packages.tex}

\makeatletter
\@ifclasswith{article}{anonymous}
{%
    \usepackage{comment}
    
    \usepackage[true]{anonymous-acm}
    \renewcommand*{\textanon}[2]{\ifstrequal{}{#2}{(text removed for peer review)}{#2}}
}{%
    \usepackage[false]{anonymous-acm} 
}
\makeatother

\usepackage{graphicx} 
\usepackage{array,siunitx}
\usepackage{textgreek}
\usepackage{tabularray}
\UseTblrLibrary{booktabs} 
\UseTblrLibrary{siunitx} 
\UseTblrLibrary{varwidth} 
\usepackage[tableposition = top]{caption} 
\usepackage{booktabs} 
\usepackage[l3]{csvsimple}
\usepackage{hyperref}
\hypersetup{
    colorlinks = true,
    allcolors = {blue}
}
\usepackage{subcaption}
\usepackage[export]{adjustbox}
\usepackage{wrapfig}
\usepackage{verbatim}
\usepackage{xcolor} 

\usepackage{csquotes}
\MakeOuterQuote{"}

\usepackage[nameinlink,noabbrev]{cleveref}
\usepackage{ifdraft}
\ifoptionfinal{

    \newcommand{\com}[1]{}
    
    \newcommand{\del}[1]{}
}{
    \usepackage{proofread} 
    \usepackage{setspace}
}

\usepackage{fancyvrb}
\DefineShortVerb{\|}
\newcommand\widthmodifier{0.92}

\title{Influencing Incidental Human-Robot Encounters: Expressive movement improves pedestrians' impressions of a quadruped service robot}
\date{}
\author{Anonymized for Review}

\renewcommand{\cite}{\autocite}
\begin{document}

\author{Elliott Hauser \\
 University of Texas at Austin \\
 {\underline{eah13@utexas.edu}} \\ \\
  Alekhya Kuchimanchi\\
 University of Texas at Austin  \\
 {\underline{akuchima@utexas.edu} } \\ \\
\And
 Yao-Cheng Chan \\
 University of Texas at Austin \\
 {\underline{ ycchan@utexas.edu} } \\ \\
 Hanaa Siddiqui \\
 University of Texas at Austin \\
 {\underline{ hanaasiddiqui@utexas.edu} } \\ \\
 \And
  Ruchi Bhalani \\
 University of Texas at Austin \\ 
 {\underline{ ruchi.bhalani@utexas.edu} } \\ \\
  Justin Hart\\
 University of Texas at Austin \\ 
 {\underline{ justinhart@utexas.edu } }}

\maketitle

\input{abstract}

\subsubsection*{Keywords:
human-robot encounters, quadruped robots, pedestrians, robot body language}

\section{Introduction}
\input{introduction.tex}

\section{Background}
\input{background.tex}

\section{Body Language Elicitation \& Implementation}
\input{intervention.tex}

\section{Human-Robot Encounter Study Design}
\input{study-justin.tex}

\section{Results}
\input{results.tex}

\section{Discussion}
\input{discussion.tex}

\section{Conclusion} 
\input{conclusion}

\singlespacing
\printbibliography

\end{document}

%% file: hicss-packages.tex
\usepackage[letterpaper]{geometry}
\usepackage{hicss}
\usepackage{times}
\usepackage[none]{hyphenat}
\usepackage{url}
\usepackage{latexsym}
\usepackage{indentfirst}
\usepackage{graphicx}
\graphicspath{{images/}}
\usepackage[
    style=apa,
    apamaxprtauth=6,
    doi=false,
    url=false,
  ]{biblatex}

%% file: abstract.tex
\begin{abstract}
A single mobile service robot may generate hundreds of encounters with pedestrians, yet there is little published data on the factors influencing these incidental human-robot encounters. We report the results of a between-subjects experiment (\mbox{n=222}) testing the impact of robot body language, defined as non-functional modifications to robot movement, upon incidental pedestrian encounters with a quadruped service robot in a real-world setting. 
We find that canine-inspired body language had a positive influence on participants' perceptions of the robot compared to the robot's stock movement. This effect was visible across all questions of a questionnaire on the perceptions of robots (Godspeed). We argue that body language is a promising and practical design space for improving pedestrian encounters with service robots. 
\end{abstract}

%% file: introduction.tex
Building upon under-recognized early works on human-robot interaction (HRI) during ad-hoc encounter scenarios \cite{Bergstrom2008-kq,Dondrup2014-xp,Rehm2013-wn}, human-robot interaction (HRI) researchers are recognizing incidental encounters between humans and robots as an increasingly common yet poorly-understood area within HRI \cite{Joline2020-ha,Avelino2021-cv,Hardeman2021-ki,Thunberg2020-dk,Babel2022-wv}. \citeauthor{Rosenthal-von_der_Putten2020-ev} \autocite*{Rosenthal-von_der_Putten2020-ev} have labeled the those who unexpectedly encounter robots in the course of the daily lives \textit{incidentally copresent persons} (InCoPs).

Given the demonstrated potential for agile quadrupeds to evoke fear \cite{Heller2021-pj,Sparrow2016-na}, \textcite{Yunus2021-fu} argue that HRI has an ethical duty to understand the factors influencing encounters with mobile quadruped robots. Amidst increasing research interest in InCoPs and human-robot encounters, the field of HRI is poised to produce guiding knowledge for the development and adaptation of service robots to be deployed in public, shared, and pedestrian spaces. 

To ensure this momentum continues, new methods and new kinds of studies are needed. We provide a definition of human-robot encounter studies in \cref{sec:definition} to guide research on InCoPs and the development of socially compliant robotic autonomy for unstructured, real-world environments.

This study's main contribution is an evaluation of the effects of a specific intervention, the Body Language of a quadruped robot, that demonstrates a positive impact upon self-reported InCoP experience. The intervention is evaluated in a real-world human-robot encounter. A premise of our approach is that the future adoption of quadruped service robots in pedestrian spaces will occasion many incidental encounters between service robots and non-users. In this study, we thus chose to investigate how body language implemented on the Boston Dynamics Spot quadruped robot platform impacts InCoPs experience and perception of the robot. Body language in this context is defined as the expressive character of robotic locomotion not required for the performance of an activity \cite[see][]{Venture2019-vu}. Body language is an important potential design intervention, since it is implementable during the performance of a range of other duties robots may perform that lead to co-presence with humans. 
Our results demonstrate that relatively simple changes to service robot body language can have a positive impact upon human-robot encounters in a pedestrian context. 
Body language interventions could be a particularly promising intervention for existing and future deployments of mobile robots in public settings where incidental human-robot encounters are most common.

%% file: background.tex
\label{sec:background}

Mobile service robotics research advances have enabled autonomous or semi-autonomous robots to successfully navigate complex, highly populated spaces to perform tasks for users. Participants' initial reactions to these robots, which prominently include curiosity, evolve substantially over time. This well-known HRI phenomenon, the "novelty effect", is a complicating factor for longitudinal social impacts of real-world deployments \cite{Hart2022-km}. A recent report by the Knight Foundation on multi-year robot delivery programs in four US cities found that "While curiosity was a common initial reaction, the responses were mixed and others expressed a disinterest in seeing robots on city streets. During the first
virtual community meeting in Pittsburgh, many people voiced concerns about the deployment of robots and expressed displeasure with the pilot." \cite{Howell2022-ac}. 
Regardless of robots' benefits to the customers and businesses that use them, negative pedestrian perceptions risk degrading the social function of public spaces and ultimately eroding support for mobile robotics. 

The documented transition from curiosity to concern is a critical motivation and framing for studies of human-robot encounters. Exogenous factors like news coverage and public perception of specific robotic platforms in, for instance, policing \cite{Yunus2021-fu}, may play a role in how public perceptions shift over time. Regardless, we suggest that HRI researchers can and should focus on endogenous factors of public concern. A justification for this is the observation that, during the periods cited above where public perception shifted from curiosity to concern, large numbers of human-robot encounters have occurred. Researchers know little about such encounters and even less about how to influence them, seriously hampering the field's ability to ensure positive experiences during incidental human-robot encounters.  

HRI research in encounter scenarios has, however, begun to reveal the specific nature of pedestrian concerns. Hardeman \cite*{Hardeman2021-ki} conducted field research on 28 InCoPs who walked past mechanomorphic delivery robots. Interview and survey results revealed several concerns, including privacy, employment (job loss), and collision. Babel et al. \cite*{Babel2022-wv} deployed a mechanomorphic robot in a busy train station. Observations and interviews with passersby showed concerns about collision, job loss, and the inconvenience caused by the robot's lack of communication capabilities.

Initial research has suggested several ways to approach this challenge. Moesgaard et al. \cite{Moesgaard2022-jp} suggest that field studies, in contrast with traditional lab-based studies, hold the key to understanding human-robot encounters and InCoP experience. Their ethnographic results suggest that designs of service robots in pedestrian settings must study robot capabilities and pedestrian perceptions of robots together. Avelino et al. \cite*{Avelino2021-cv} explored the literature on social robots' greeting abilities in encountering new users in a social context. This work underlines the importance of first impressions to InCoP experience during encounters that are often very brief. To date, however, few studies have directly evaluated specific interventions' ability to improve InCoP experince.

Traditional HRI research has demonstrated that emulating aspects of canine behavior can improve perceptions of a range of human-robot interactions. Commercially available companionship robots such as AIBO or Golden Pup have been shown to elicit positive reactions in children \cite{Row2020-fm}, the elderly \cite{Ihamaki2021-aq}, and long-term owners of the robot \cite{Kertesz2017-ix}. Some studies suggest that human-dog interactions are less effective models for social robot interactions \cite{Feil-Seifer2014-rx,Kerepesi06} and are not directly transferable to service applications or the incidental encounter context. Other evidence, however, shows that canine characteristics are remarkably versatile robotic interventions: a tail has been shown to enhance perceptions of commercially available robotic vacuum cleaners \cite{Singh2013-yu}. 
There is evidence that these effects apply across interaction modalities, further indicating the robustness of these effects. Interactive canine behaviors have positively impacted human interactions with virtual robots experienced via AR \cite{Norouzi2019-nj}. 

Given the clear gaps in evaluating specific interventions in incidental encounters and the demonstrated effectiveness of canine-inspired behavior, there is ample justification for research on the influence of quadruped body language on incidental human-robot encounters.

%% file: intervention.tex
We developed a quadruped body language intervention designed to preserve theoretical feasibility in real-world service robot deployments by intervening in software only, using existing platform capabilities. We used an elicitation survey to guide design of canine-inspired body language indicative of positive affective states implementable on the Boston Dynamics Spot without hardware modification. We implemented these as a Body Language Vignette using the Boston Dynamics Spot. A Control Vignette taking the same amount of time but without these behaviors was developed as well. The Vignettes' repeatability ensured that the robot's behavior was comparable within each of of the two groups of our inter-participant experimental design (described further below).

\subsection{Selection of Body Language Elements}
To inform our selection of canine-inspired body language for development, we elicited free-text responses from online survey respondents (\mbox{n=57}) about canine behaviors that they perceived as connected to six affective states. Then, after watching a short video of media coverage showing the Boston Dynamics Spot moving, we elicited suggestions of improvements to the robot's behaviors. We utilized qualitative coding to analyze the relative frequency and overall patterns in elicited behaviors. The final behaviors were selected by the authors based on their suitability for implementation without hardware modification and the observability requirement of human-robot encounters. 

The behaviors selected for implementation as canine-inspired quadrupedal robot body language in this study are tail wagging, play bow, sitting, walking in circles, and chasing its tail. In addition to selecting these behaviors, the elicitation survey results were used to inform implementation. Suggestions judged to be too ambiguous to be implemented directly, such as "move in a calm and relaxed manner", smoother movement, and varied walking speeds were incorporated across the relevant behaviors as they were refined.

\subsection{Implementation}
\input{figures/movement_images}

To create the Body Language Vignette, each of the chosen behaviors was hand-coded as a motion approximating each selected canine behavior.
Each motion is defined as a set of trajectory points that the robot follows, using a custom open-licensed C++ API for the Boston Dynamics Spot.\footnote{The API can be found at \textanon{\url{https://github.com/ut-amrl/spot_cpp}}{(link to public GitHub repository with code used in this and other studies)}. The development team for the API includes \textanon{Nathaniel-Nemenzo, Mateusz Kozlowski, Swathi Mannem, Daksh Dua, Shikhar Gupta, Geethika Hemkumar, Maxwell Svetlik, Parth Chonkar, Marika Murphy, Joydeep Biswas, and Justin Hart.}{(list of contributors including one or more authors removed for review}}  

Motions are defined with respect to a neutral body posture, with the robot's torso parallel to the floor and its legs slightly bent. The motions are periodically and angles with respect to the center of the torso. 
There are two basic trajectory types. One is based on roll, pitch, and yaw coordinates with respect to the robot's body in a neutral posture, with the robot moving through those coordinates, but not otherwise walking. The other trajectory type is deltas with respect to the robot's current position, specified as a velocity (back-to-front along the x-axis, or left-to-right along the y-axis) and a rotation through the robot's yaw angle. Images of all motions are provided in \ref{fig:tailwag}.
In the wagging motion (\ref{fig:tailwag}), the robot rocks synchronously across its roll and yaw axes, between the extremes of $\frac{-\pi}{16}$ to $\frac{\pi}{16}$, and $\frac{-\pi}{8}$ to $\frac{\pi}{8}$ radians off of the neutral pose. The robot tilts to the right, returns to the center, tilts to the left, returns to the center, and continues in a smoothly-interpolated motion. 
%
The play bow motion (\ref{fig:play-bow}) approximates a dog ``bowing'' during play (a canine invitation to engage in play). The robot tilts its torso entirely forward to the edge of its ability, at $\frac{3\pi}{14}$ radians, and then back.
%
The sit motion is the opposite of play bow, at a less severe angle of $\frac{\pi}{7}$ See Figure \ref{fig:sit}.
%
To walk in a circle, the robot follows uniform linearly-interpolated waypoints placed at $1.5$ radians per second of rotation about the yaw axis and $2$ meters per second of forward motion. See Figure \ref{fig:sit}.
%
The ``spin'' motion mimics a dog chasing its tail, which was one of the popular survey responses. Spinning is the motion closest to tail-chasing that the robot can make as it does not possess a spine and cannot twist to mimic actual tail-chasing. To execute this motion, the robot moves through waypoints placed at $1$ radian per second, similar to the ``Walk in Circle'' motion, but with no forward motion component. See Figure \ref{fig:spin}.


In the \textit{Body Language Vignette}, the robot repeatedly performed all selected behaviors. In the \textit{Control Vignette,} the robot walks in a semicircle, then continues forward for a few steps before entering into another semicircular pattern. The non-canine walking behavior loops four times before the robot comes to a stop. The location and duration of both vignettes are the same. Body language is thus the only intervention across encounter conditions. Videos of both vignettes are available online.\footnote{\textanon{Body Language Vignette: \url{https://youtu.be/R9vWQJpmh-0}\\Control Vignette: \url{https://youtu.be/f2sfL1cJ4PE}}{(For publication we plan to provide YouTube URLs of existing demonstration videos of each Vignette here. Since these would de-anonymize us, we instead provide \href{https://drive.google.com/file/d/1eADVXheXrecm10SBQU60PXWgqNntELmC/view?usp=sharing}{this Drive link to an anonymized review video}} of both Vignettes.)}


%% file: figures/movement_images.tex
\begin{figure}[tb]
    \centering
    \begin{subfigure}[b]{0.960\columnwidth}   
        \centering
        \includegraphics[width=\linewidth]{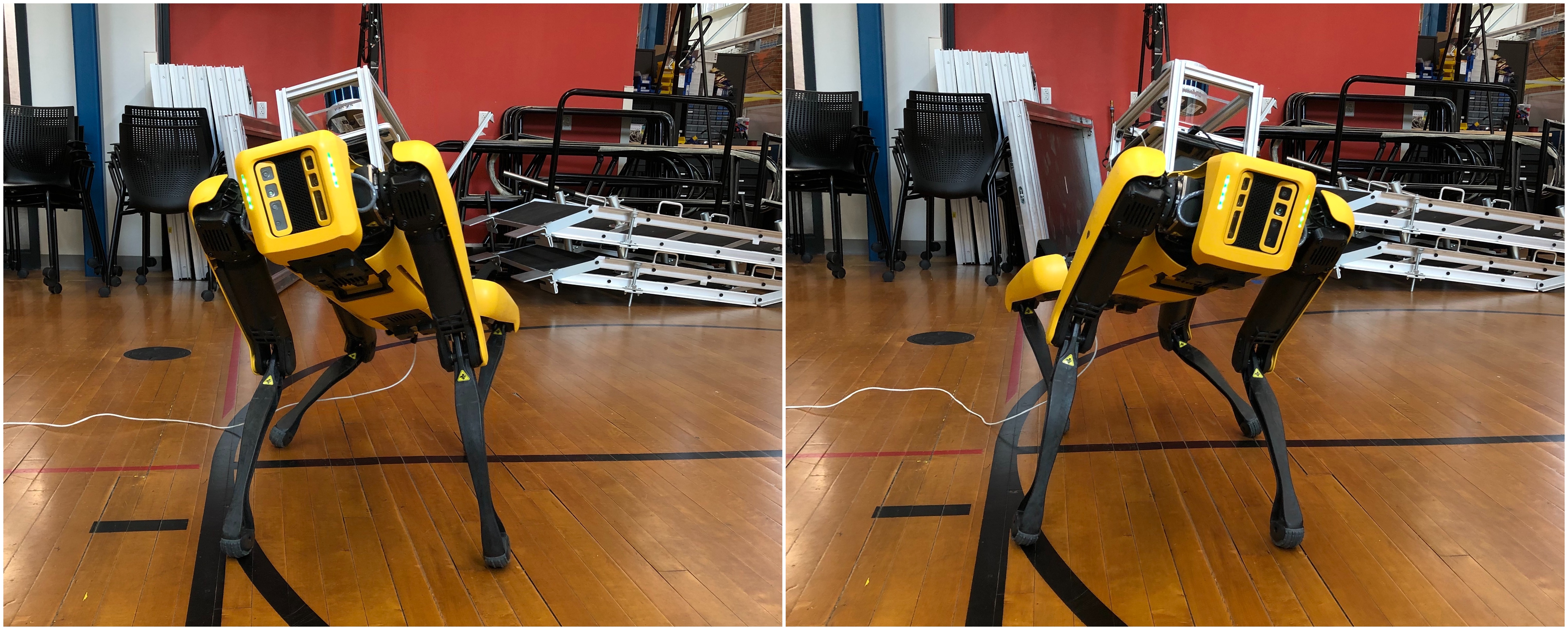}
        \caption[]%
        {Wagging}
        \label{fig:tailwag}
    \end{subfigure}
    \begin{subfigure}[b]{0.475\columnwidth} 
        \centering
        \includegraphics[width=\linewidth]{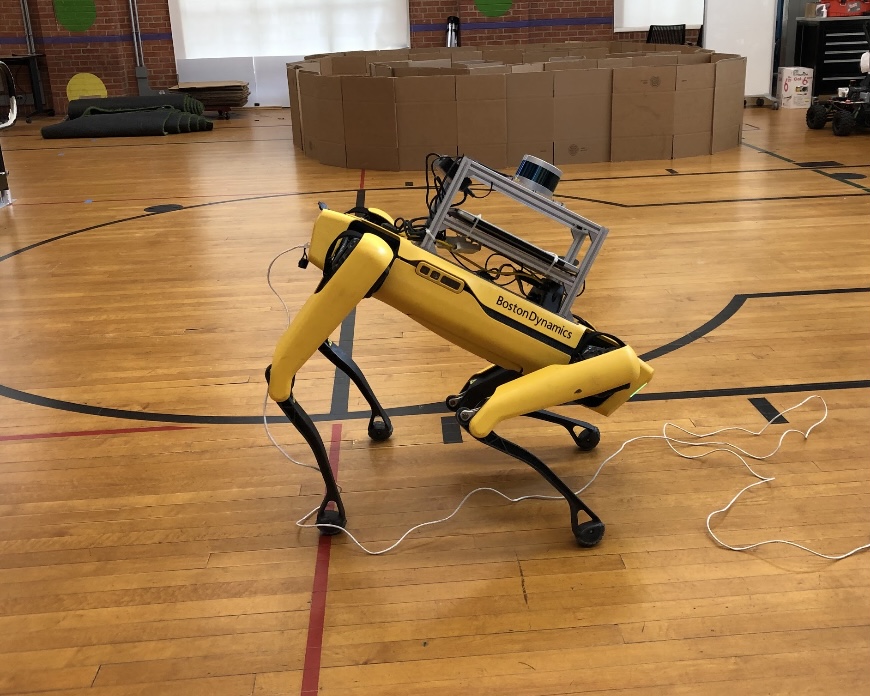}
        \caption[]%
        {Play bow}
        \label{fig:play-bow}
    \end{subfigure}
    \begin{subfigure}[b]{0.475\columnwidth} 
        \centering
        \includegraphics[width=\linewidth]{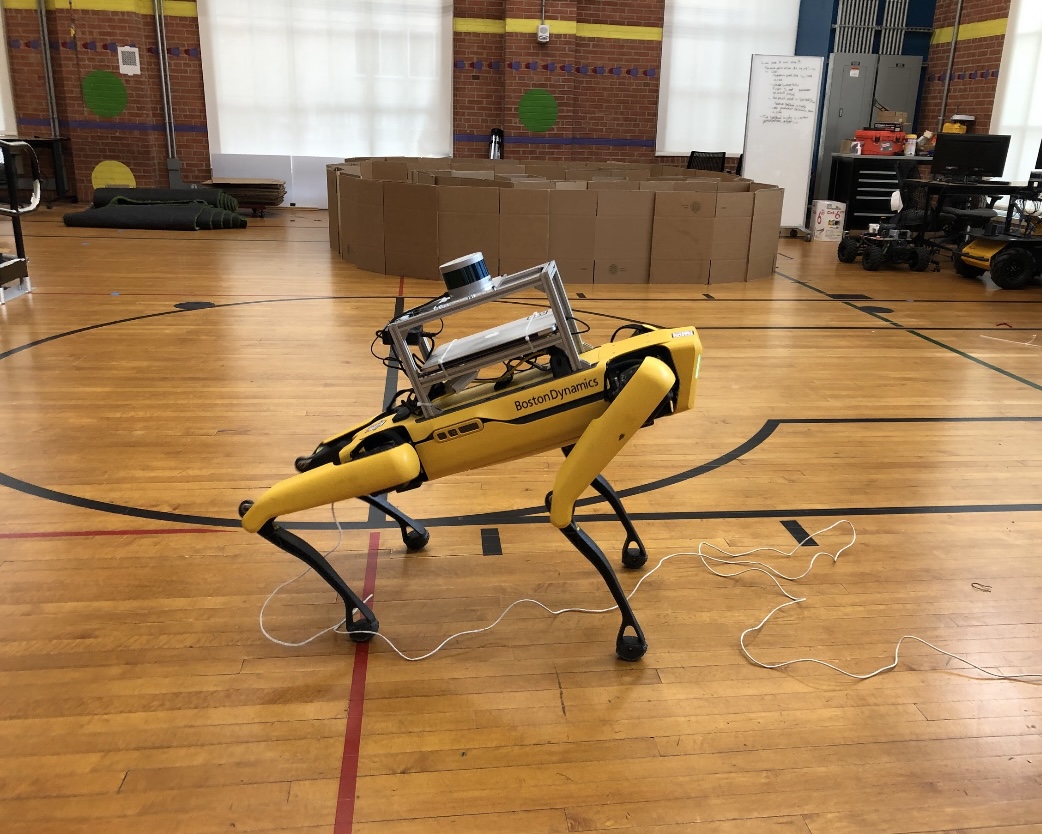}
        \caption[]%
        {Sit}
        \label{fig:sit}
    \end{subfigure}
    \begin{subfigure}[b]{0.475\columnwidth}  
        \centering 
        \includegraphics[width=\linewidth]{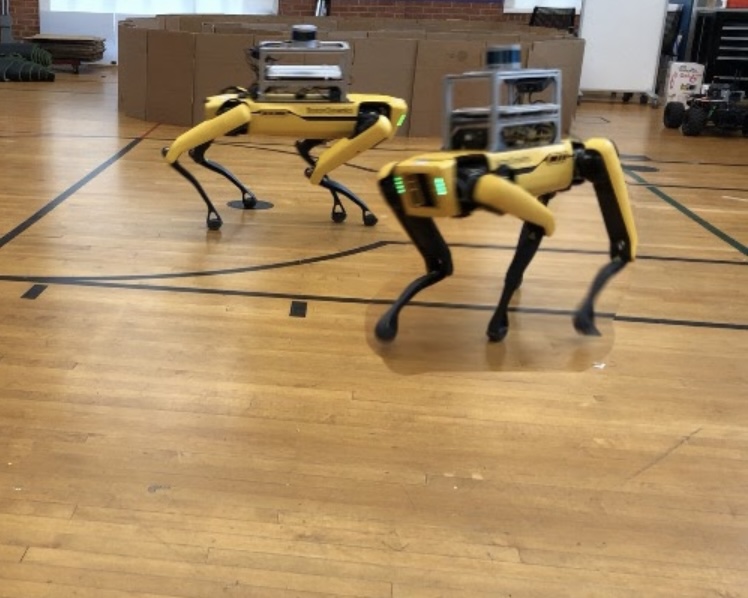}
        \caption[]%
        {Walk in circle}
        \label{fig:circle}
    \end{subfigure}
    \begin{subfigure}[b]{0.475\columnwidth}   
        \centering
        \includegraphics[width=\linewidth]{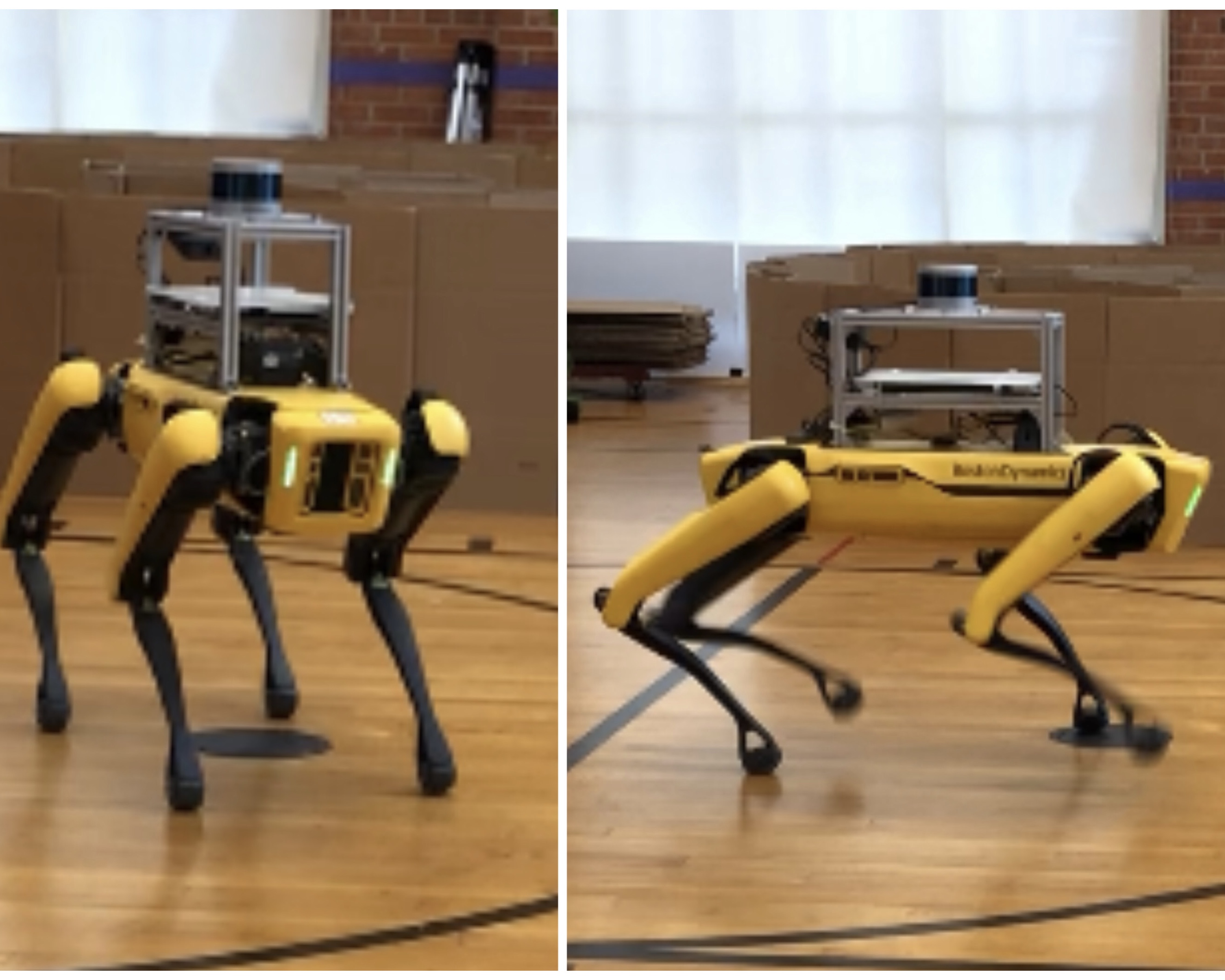}
        \caption[]%
        {Spin}
        \label{fig:spin}
    \end{subfigure}
    \caption[]%
    {Images of the robot performing the  behaviors that comprise the Body Language Vignette.}
    \label{movement-pics}
\end{figure}

%% file: study-justin.tex
This section provides a definition of human-robot encounters, a human-robot encounters study, a rationale for the design of our intervention, and a description of our survey procedures. Limitations of the design are discussed in \cref{limitations}.

\subsection{Definitions}
\label{sec:definition}
We describe this work with term \textit{human-robot encounter study} rather than the more familiar \textit{user study} to indicate the difference in focus from traditional HRI study designs \cite{Pentzold2022-hh,Joline2020-ha,Rehm2013-wn,Dondrup2014-xp}. This terminological innovation is in part inspired by \citeauthor{Rosenthal-von_der_Putten2020-ev}, who noted that there was no commonly accepted term for the human in an incidental human-robot encounter. They propose the term \textit{incidentally co-present person (InCoP)}  in \cite{Rosenthal-von_der_Putten2020-ev}, which we adopt here and incorporate into our definitions. 

We define a \textit{human-robot encounter} as any recognized, unplanned, incidental spatiotemporal co-location of human and robot. The recognition dimension indicates that at least one human or robot must \textit{perceive} the presence of the other. The unplanned dimension excludes instances where the human has planned to encounter a robot, such as when receiving a delivery order or operating the robot. Spatiotemporal co-location is a permissive view of an encounter's extent: proxemic thresholds are avoided in favor of the inherent limitations of perception. A co-present human expecting a robot cannot incidentally encounter it and is by definition not an InCoP in the context of that encounter.

The term \textit{human-robot encounter study} is most appropriate when \textit{InCoPs} are the \textit{subject} of a study set in or designed to replicate the conditions of an incidental encounter as defined above. This term helps elevate the incidental encounter as a distinct spatiotemporal context in need of focused HRI research. Finally, it helps disambiguate the InCoPs from an end user or beneficiary of a service robot. Traditional HRI studies' focus on direct interaction between humans and robots \cite{Moesgaard2022-jp} may feature in human-robot encounter studies insofar as it is a factor in InCoP-robot encounters. Our human-robot encounter study minimizes such factors in an effort, limiting our intervention to robot body language.

\input{figures/godspeedTable.tex}

\subsection{Rationale}
\label{sec:studyrationale}

We chose a vignette-based implementation of body language to enhance its isolation as a factor in our intervention in the human-robot encounter in this study. Each Vignette was reproducible and was operated near-continuously during experiments. This meant that participants who encountered a given Vignette experienced very similar robot behavior as other participants in that study condition. This allowed more comparability across conditions and participants.  

The Vignettes were not obviously connected to a specific service task, which was a drawback to this design. We believe that the benefits of this design outweigh the drawbacks: \textit{InCoPs} may not know what task a service robot is engaged in, or even whether it is engaged in a task during an encounter. A realistic simulation of a service task would have required the robot to navigate a wider area, making survey recruitment more logistically difficult. The Vignette design eased in-situ participant recruitment (discussed below), balancing logistical feasibility with ecological validity.


The outdoor pedestrian environment selected for the study allowed potential participants to perceive the robot from a variety of distances, mimicking the conditions under which encounters with service robots might occur. Compared with indoor settings, this choice increased the number of potential study participants.


We chose an in-person study design, even though a video-based survey design could have solicited a larger participant pool in less time. Some evidence suggests that video is equivalent in many respects to in vivo experience \cite{Woods2006-dn,Babel2021-xf}, but in all cases there were observed differences. We believe in-person modalities have the highest ecological validity in human-robot encounter studies, especially with relatively rare platforms like the Spot that nonetheless are beginning to receive wide media coverage. The incidental nature of pedestrian encounters is extremely difficult to replicate via video. Until research can establish the relative validity of virtual encounter modalities like video, in-person designs are preferred for human-robot encounter studies. 

\subsection{Survey Methodology}
\label{human_survey_methodology}
The questionnaire used was derived from the Godspeed Questionnaire \cite{bartneck2009measurement}, which measures perceptions of robots on scales of Anthropomorphism, Animacy, Likeability, Perceived Intelligence, and Perceived Safety. We determined that the questionnaire must be shortened to avoid survey abandonment, given the in vivo robot operation and in situ recruitment, which required participants to use their mobile devices for survey completion.  \citeauthor{bartneck2009measurement} suggests researchers modify it to meet the needs of a study, as we did here. 

We determined that questions implying human characteristics were unlikely to yield meaningful data in comparison to others in the same category, and modified them to reference dog-like characteristics where appropriate, or removed them. This included replacing "Humanlike" with "Doglike" and replacing the non-participant-facing "Anthropomorphism" category with "Cynomorphism". The comparisons presented in the questionnaire are displayed in Table \ref{table:godspeedTable} under their respective categorical descriptions (which were not displayed to participants).

\input{figures/site.tex}

The robot ran the Body Language Vignette and the Control Vignette on two separate weekdays, at a busy pedestrian intersection on \textanon{the University of Texas at Austin} campus in March 2022. Classes were in session at the time, and the robot was operated during the busiest part of the day. 

The study consisted of a flat, concrete square area, adjacent to two roads. Two of the sides were accessible by foot (North and East sides) while the other sides were blocked with low barriers providing partial visibility. Spot performed the behaviors relative to the center of the area. The robot was positioned at the corner of a building to increase the likelihood of participation and that encounters would occur at a relatively close range. The experimental arrangement is detailed in \cref{fig:site}.

On both days, QR codes linking to an online survey were displayed nearby. A light pole on the southeast corner provided a prominent place to place the QR code for the survey.
Participants who encountered the vignettes voluntarily used the QR codes to complete the questionnaire on their mobile devices. Researchers did not encourage bystander participation or recruit participants verbally.
The QR codes were removed after the study sessions, and the team verified that all survey responses were received during the times the robot was being operated, ensuring that all responses were genuine. The participants signed an informed consent prior to answering the survey. This study is approved by the Institute Review Board of \textanon{UT Austin}{}.

%% file: figures/godspeedTable.tex
\begin{table}[htb]
\centering
\begin{tabular}{|c|} 
 \hline
 \textbf{Anthropomorphism (Cynomorphism)} \\ \hline
 Machinelike vs Doglike\\ 
 Fake vs Natural\\ 
Unconscious vs Conscious \\ \
Artificial vs Lifelike  \\ 
Moving Rigidly vs Moving Elegantly \\\hline
\textbf{Animacy} \\ \hline
Inert vs Interactive \\ 
Apathetic vs Responsive \\ \hline
\textbf{Likability } \\ \hline
Dislike vs Like \\
Unfriendly vs Friendly \\ 
Unpleasant vs Pleasant \\	
Awful vs Nice \\ \hline	
\textbf{Perceived Intelligence} \\ \hline
Incompetent vs Competent \\ 
Ignorant vs Knowledgeable \\ \hline

\end{tabular}
\caption{The abridged Godspeed Questionnaire  used in this study, presented to participants as the bipolar questions opposing these terms. The full questionnaire is available in \protect\cite[79]{bartneck2009measurement}. 
}
\label{table:godspeedTable}
\end{table}

%% file: figures/site.tex
\begin{figure}[bt]
\centering
  \includegraphics[width=\widthmodifier\columnwidth]{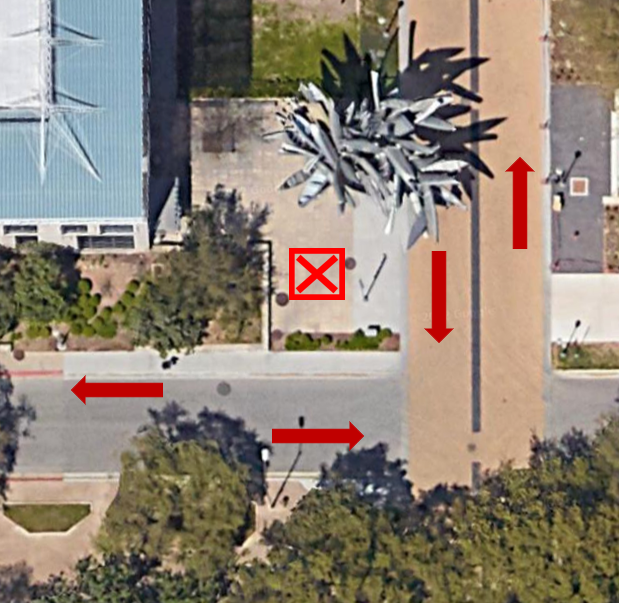}
  \caption{Overview of the experiment environs. The red box indicates the location where the robot was operated, at the intersection of two large pedestrian streets. Red arrows indicate the typical flows of bidirectional traffic.}
  \label{fig:site}
\end{figure}

%% file: results.tex
\label{resultsSection}
\input{figures/godspeedcategories.tex}
\input{figures/godspeedfig}

A total of 222 people responded to the survey: $112$ saw the Body Language vignette (the robot performed the selected behaviors repeatedly) and $110$ saw the control vignette (the robot walked in a semicircle pattern repeatedly). The participant pool consisted of 202 students, 6 faculty members, and 14 individuals with non-university affiliations.

The same survey was distributed to all participants and one way ANOVA was used to compare the survey results of both vignettes. The results in \cref{tab:godspeedcategories} show a significant difference in Cynomorphism ($F_{1,220}$=$4.10$, $p$=$0.04$) and Animacy ($F_{1,220}$=$6.18$, $p$=$0.01$), a marginally significant difference in Likeability ($F_{1,220}$=$3.21$, $p$=$0.07$), and no significant difference in Perceived Intelligence ($F_{1,220}$=$0.18$, $p$=$0.67$). 

The aggregated results of these high-level dimensions of Godspeed indicate that body language is an effective way to elicit a sense of cynomorphism and animacy in a quadruped robot, as well as an easy manipulation to make a quadruped robot more likable. The non-significant difference in Perceived intelligence is foreseeable due to the vignettes' disconnection from an actual task where robots' intelligence and abilities play a major role. 


Participants' perceptions also significantly improved on several metrics in the body language intervention versus the control condition. First, the participants found the robot to be more dog-like in the Body Language Vignette. This result suggests that the implemented behaviors reasonably evoked the canine behaviors elicited by our formative survey. Secondly, despite the robot not actually interacting with them or responding to them, participants perceive the robot as more responsive 
 and more friendly 
 when performing the Body Language vignette. Lastly, the participants who encountered the Body Language Vignette also perceived the robot as more conscious. In fact, though not statistically significant on each individual metric, participants rated the robot more favorably on every single scale in Godspeed category that was measured. 


The modified Godspeed Questionnaire preserved the original question groupings where appropriate, enabling the combination of individual questions to derive an overall perception of Cynomorphism (modified from the original Anthropomorphism), Animacy, Likeability, and Perceived Intelligence. Tests of statistical significance via ANOVA of these findings, as well as a combined measure of all questions, are given in \cref{tab:godspeedcategories}. A visualization of results is provided in \cref{fig:godspeedfig}.

%% file: figures/godspeedcategories.tex
\begin{table*}[htb]
    \caption{Analysis of individual survey results including Godspeed category aggregates. 
    }
    \centering
    \label{tab:godspeedcategories}
    \footnotesize
    \NewColumnType{F}[1][]{Q[si={table-format={<<}2.2,round-mode = places,round-precision = 2,round-minimum = 0.01,text-family-to-math = true , text-series-to-math = true,#1},c]}
    \csvreader[
        head to column names,
        collect data,
        generic collected table = tblr,
        generic table options = {{
            row{1-2} = {guard},
            row{3,9,12,17,20} = {font={\bfseries}},
            hline{2,3-4,9-10,12-13,17-18,20-21} = {1pt, solid},
            colspec={Q[r,f]X[1,c]@{}X[-1,c]X[1,c]@{}X[-1,c]FF},
            cell{1}{2,4} = {r=1,c=2}{c},
            cell{1}{1,6,7} = {r=2}{f},
            width=.8\linewidth,
        }},
        table head = { & \bfseries Body Language &&\bfseries Control&&{{{$F$ value}}}& {{{$p$ value}}} \\
        &Mean&\textsigma&Mean&\textsigma&&\\ }
    ]{data/godspeedstats.csv}{}{%
\csvexpval\name & 
    \tablenum[table-format=1.3,round-mode = places,round-precision = 2,text-family-to-math = true , text-series-to-math = true]{\blmean} 
        & \tablenum[table-format=1.2,round-mode = places,round-precision = 2,text-family-to-math = true , text-series-to-math = true]{\blsd}
        & \tablenum[table-format=1.3,round-mode = places,round-precision = 2,text-family-to-math = true , text-series-to-math = true]{\cmean}
        & \tablenum[table-format=1.2,round-mode = places,round-precision = 2,text-family-to-math = true , text-series-to-math = true]{\csd}
        & \tablenum{\fvalue} 
        & \tablenum{\pvalue}}
\end{table*}

%% file: figures/godspeedfig.tex
\begin{figure*}[htb]
    \centering
    \includegraphics[width=\widthmodifier\linewidth]{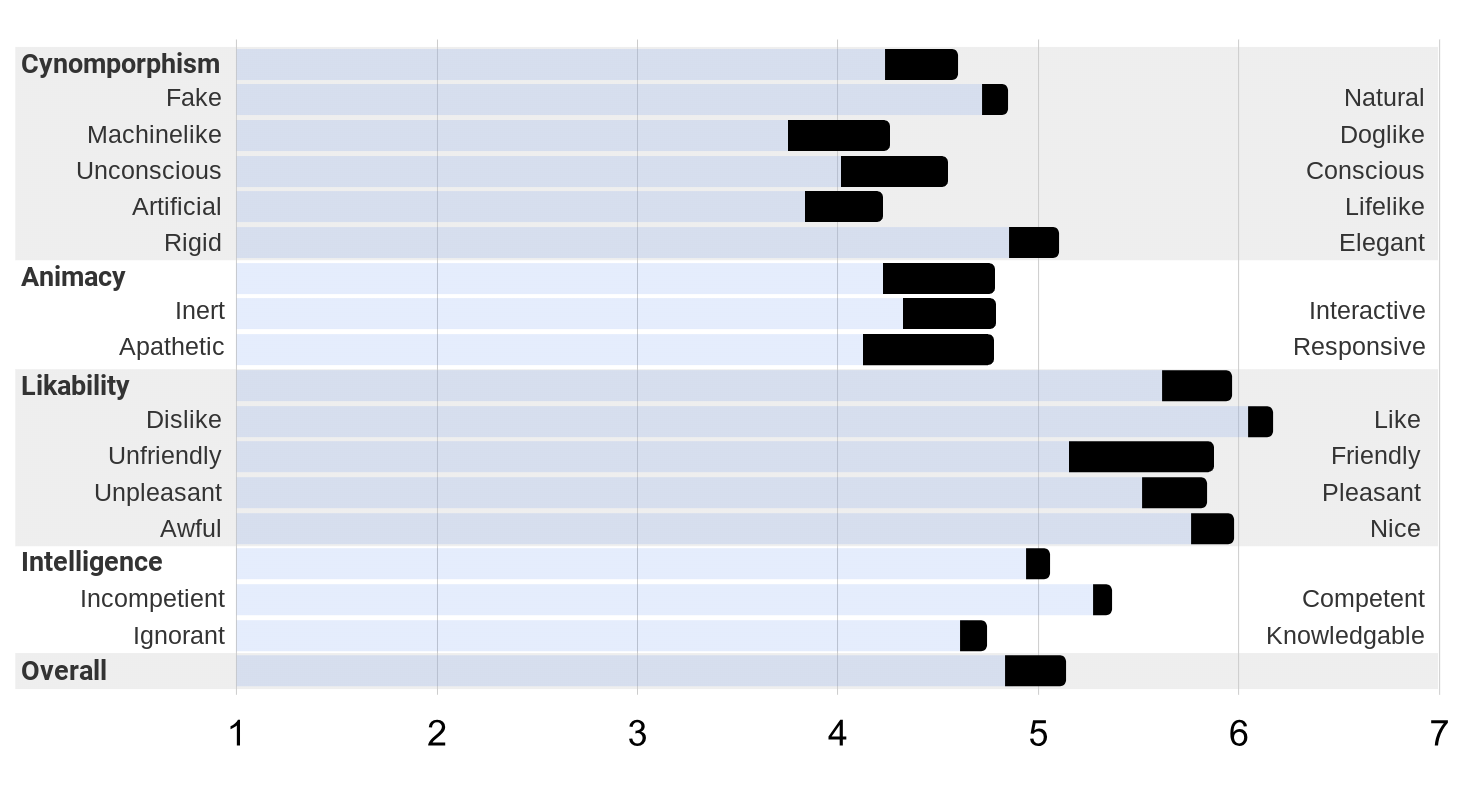}
    \caption{Visualization of survey results by individual Godspeed question and Godspeed category. The light blue bar represents mean responses for the Control condition. The black bar represents the change in mean reported between the Body Language and Control Vignette conditions in this inter-participant design. Category-level results average the responses for each constituent question. See \cref{tab:godspeedcategories} for significance calculations.}
    \label{fig:godspeedfig}
\end{figure*}

%% file: discussion.tex
Together, these results indicate that body language interventions on quadruped service robots using existing hardware and capabilities can be leveraged to positively improve {InCoP} experiences of robot encounters.

The Body Language vignette was rated significantly higher in Cynomorphism and Animacy scales than the Control vignette, with a marginally significantly higher Likeability rating. 
This supports our hypothesis that {InCoP} experience can be positively impacted by non-functional expressive motion.
There was no difference in the Perceived Intelligence ratings between the two conditions. This indicates that other kinds of interventions are required for InCoPs to perceive autonomous robots as more cognitively capable.

Given our findings, we discuss three promising areas for future research: explicit indications of service robot role, the types and impacts of encounter context, and quadruped-specific hardware for affective expression.

\subsection{Indicators of Service Robot Role}
Drawing inspiration from service animals, quadruped service robots could bear idiomatic indicators of their service role, such as a vest, leash, or harness \textanon{\cite{hauser2023s,Chan2023-dd}}{(Author(s) 2023, Author(s) Forthcoming)}. 
In addition to a deeper investigation of apparent robot role, human-robot encounter studies should explore other insights available from adjacent areas of sociotechnical research. For instance, the argument we made for the importance of in vivo study designs aligns well with \textit{tactical urbanism} that \textcite{Joshi2019-gs} used to engage communities in the appropriate development of community or municipal robotics deployments. Collaborations of this form could let HRI play a larger role in shaping the future of urban pedestrian experience.

\subsection{Modulation of Effects by Encounter Context}
With more studies on human-robot encounters and evidence for the effectiveness of different kinds of interventions in different kinds of contexts, it will become necessary to develop an understanding of the properties and interactions between intervention and context to guide their use. Considering the proxemic characteristics of human-robot encounters, they may often happen at a distance. Visible signifiers such as vests \textanon{\cite{hauser2023s}} may have a longer effective distance than body language as presented here or other forms of non-functional expressive motion such as simulated breathing \cite{terziouglu2020designing}. The disparate outcomes of cross-cultural studies of HRI \cite{Strait2020-hl,Winkle2022-ij} suggest that cultural contexts of encounter likely moderate human-robot encounters. We expect that the strength and nature of contextual moderation will vary across and within intervention types.  Finally, individual factors, such as visual disabilities, will play an important role as well. 

\subsection{Affective Hardware for Quadruped Service Robots}
\label{sec:tail}
Future work should explore the possibilities available without our self-imposed restriction to use a platform's existing hardware. Given the prevalence of tail-related behavior in our formative elicitation survey of expected canine behavior, we believe it likely that tail-like hardware could enable effective body language modulation for quadruped robots that is easily visible at a range of distances during human-robot encounters. We have found no published research concerning the addition of robotic tails to quadruped service robots. The most directly relevant study on the addition of tails as affective indicator, from 2013, found tails added to an iRobot-based platform to be consistently interpreted as affective indicators by participants (n=20) \cite{Singh2013-yu}. This is evidence that affective hardware may positively impact the perception of mobile service robots and is under-explored.

\subsection{Limitations}
\label{limitations}
Our study design had several limitations that can be mitigated in future work building from these results. Firstly, News coverage and adoption of specific robotic platforms in, for instance, policing \cite{Yunus2021-fu}, is an example of exogenous, pre-encounter factors not measured in this study. It would have been ideal to characterize participants' experience with robots to locate them within a timeline of exposure to other robots. We do not collect psychometric data that could have let us determine the degree to which such metrics contribute to participant experience. The recently proposed AMPH instrument \cite{Damholdt2020-el}, designed to be paired with the Godspeed questionnaire as a profile of participant characteristics, is a simple psychometric instrument that could augment experimental protocols with knowledge of participant characteristics. Using metrics from the Godspeed questionnaire as the main evaluation is appropriate in the present study but may not be comprehensive for understanding human-robot encounters. The Godspeed questionnaire's robot platform focus may not be an ideal operationalization of InCoP experience of a robot encounter. Other scales, such as the Perceived Social Intelligence scales \cite{Barchard2020-vp}, should be investigated for use in human-robot encounters research, ideally including Godspeed cross-validation. 

Secondly, in the Body Language vignette, the robot performs all the behaviors repeatedly; With the in vivo robot operation and in situ survey completion, we can not control what the participants (live pedestrians) actually saw. Therefore, we treat the vignettes as units of analysis, and the survey data of the Body Language vignette is the participants' general perceptions of the selected robot behaviors. While this is appropriate for this study's goal of establishing the presence of an effect, it does not provide information on the relative contribution of each selected behavior on participant experience. In addition, the vignettes are not directly implementable in service tasks where different robot behaviors may be required or favored. Instead, our results are evidence that the labor-intensive process of determining which body language might be appropriate for a given service or deployment context can have a positive impact on encounter experience.

%% file: conclusion.tex
This study is, to our knowledge, the first to directly intervene in pedestrian experiences of incidental encounters with quadruped service robots through body language. We found that participant perceptions of the encounter were positively influenced by software-only changes to a quadruped service robot's body language, defined as expressive, non-functional characteristics of the movement. Participants viewed the quadruped performing simple body language movements as more friendly, responsive, doglike, and conscious than the control, which walked in a circular pattern in the same location on a different day. Additional contributions include a definition of human-robot encounter studies and a description and rationale behind our methodological innovations. It contributes to a burgeoning literature on the experiences of InCoPs during robot encounters \cite{Moesgaard2022-jp,Thunberg2020-dk,Rosenthal-von_der_Putten2020-ev}.

Our results and the relative simplicity of body language as an intervention suggest that body language is a promising modality for positively impacting {InCoP} encounters with mobile service robots. While the expressivity and implementation details of software-only body language will vary widely by platform, it has the distinct benefit of feasibility for existing deployments and research deployments alike. 

Future research should identify how different spatiotemporal contexts, robotic platforms, and service robot tasks interact with {InCoP} experiences during unexpected incidental encounters with service robots. Finally, the complexity of human-robot encounters and growing literature demand new theoretical models. Such models will have the potential to both shape and be shaped by the empirical results of encounter-focused studies, contribute new insights to other aspects of HRI, and enable new and more successful applications of service robots in shared spaces like the pedestrian setting studied here.